# Semantic Embeddings of Chemical Elements for Enhanced Materials Inference and Discovery

Yunze Jia[1,#], Yuehui Xian[1,#], Yangyang Xu[2,#], Pengfei Dang[1], Xiangdong Ding[1], Jun Sun[1], Yumei Zhou[1,*], Dezhen Xue[1,*]

1 *State Key Laboratory for Mechanical Behavior of Materials, Xi'an Jiaotong University, Xi'an 710049, PR China*

*2 MOE Key Laboratory for Nonequilibrium Synthesis and Modulation of Condensed Matter, School of Physics, Xi'an Jiaotong University, Xi'an 710049, PR China*

*# Contributed equally.*

## Abstract

We present a framework for generating semantic embeddings of chemical elements to advance alloy inference and discovery. This framework leverages ElementBERT, a domain-specific BERT-based natural language processing model trained on 1.29 million abstracts of alloy-related scientific papers, to capture latent knowledge specific to alloys. These semantic embeddings serve as robust elemental descriptors, consistently outperforming traditional empirical descriptors across multiple downstream tasks, including predicting mechanical and transformation properties, classifying phase structures, and optimizing materials properties via Bayesian optimization. Applications to titanium alloys, high-entropy alloys, and shape memory alloys demonstrate up to 23% improvements in prediction accuracy. Notably, machine learning models based on semantic embeddings exhibit superior performance on newly published alloy data outside the training corpus, indicating promising generalization capabilities for emerging alloy compositions. Our results show that ElementBERT surpasses general-purpose BERT variants by encoding specialized alloy knowledge. By bridging contextual insights from scientific literature with quantitative inference, our framework accelerates the discovery and optimization of advanced alloys, with potential for extension to other material classes.

*Correspondence: Y. Zhou (zhouyumei@xjtu.edu.cn); D. Xue (xuedezhen@xjtu.edu.cn)

## Introduction

Chemical elements serve as the fundamental building blocks of all materials [1]. However, these elements' combinatorial nature and interactions result in an immense and complex compositional space to explore [2-4]. This vastness poses a significant challenge for traditional materials discovery approaches, which often rely on trial-and-error experimentation or computational simulations constrained by prior domain expert knowledge [5].

While Data-driven methodologies have emerged as promising tools to accelerate exploration within this space [3, 4, 6-10], they traditionally depend on structured numerical data, such as experimentally measured properties or computationally derived parameters. Effective feature engineering is often crucial for these methodologies, as the construction of representative descriptors is essential for capturing the underlying physics and chemistry required for accurate property prediction [11-13]. While effective, they overlook an invaluable source of knowledge embedded in unstructured textual formats, such as scientific literature. These texts, accumulated over decades, contain nuanced insights into material behaviors, synthesis methods, and performance characteristics that are not easily captured in numerical databases [14-16].

To bridge this gap, natural language processing (NLP) models have been employed to extract and encode semantic information from scientific texts [15, 17-22]. Models like Skip-gram and large language models (LLMs) compress textual data into numerical vector embeddings, capturing contextual meanings (e.g., the concept of "Alloys") for use in downstream machine learning tasks [23-26]. In this work, we define semantic embeddings as high-dimensional numerical representations of chemical elements learned from large-scale scientific literature by an NLP model. Unlike conventional empirical descriptors (e.g., atomic radius or electronegativity) which rely on static, tabulated physical properties, semantic embeddings capture the contextual usage and latent associations of material concepts as they appear in scientific discussions. For instance, an embedding can implicitly capture that a specific element is frequently associated with concepts like "ductility" or "high-temperature stability," effectively digitizing the domain knowledge and intuition that researchers have accumulated over decades. Since their introduction to materials science in 2019, NLP models have shown great promise in automating data extraction [27-31], predicting material properties [30, 32-36], and uncovering latent scientific knowledge [37, 38]. However, existing approaches often focus on encoding specific materials processing routes or material recipes, limiting their generalizability across material classes. A more fundamental approach involves extracting semantic embeddings of materials' building blocks, such as alloy microstructural features, crystal structures for inorganic materials [39, 40], or functional groups for organic molecules [18, 33]. These embeddings serve as generalizable representations that can enhance predictive modeling across a wide range of material systems.

Building upon this, our work distinguishes itself by establishing these semantic embeddings as a comprehensive, reusable feature library. By integrating this literature-derived knowledge base into machine learning frameworks, we can achieve a more accurate and nuanced mapping from elemental composition to material performance, transcending the limitations of traditional, fixed physicochemical descriptors. We focus on chemical elements—the most fundamental alloy building blocks—to extract their semantic embeddings and to provide a more informative alternative to traditional empirical descriptors. As shown in Figure 1a, we introduce ElementBERT, a BERT-based model trained on over 1.29 million abstracts of alloy-related literature, which reflects our focus on learning individual element-level semantic representations rather than traditional sequence-level descriptions. We specifically chose abstracts as training data because they concisely capture proposed alloy compositions alongside critical property information, mechanistic insights, and the key factors that researchers identify as performance-determining, while being universally accessible across journal publishers for reproducible model development. ElementBERT extracts semantic embeddings for chemical elements, with each embedding consisting of 384 dimensions. These element-specific descriptors are integrated with composition data through various functional forms, such as mole averaging, to construct improved input features for downstream machine-learning tasks, including regression, classification, and Bayesian optimization (BO).

The resulting ElementBERT embeddings demonstrate substantial improvements over traditional empirical descriptors across diverse alloy systems and machine learning tasks. Furthermore, beyond enhancing predictive accuracy for established alloy families, these semantic embeddings show particular promise for emerging compositions outside the original training corpus, suggesting their potential utility in guiding the discovery of novel alloy chemistries. The domain-specific training approach proves crucial, as ElementBERT consistently outperforms general-purpose language models in capturing alloy-relevant semantic relationships. ElementBERT embeddings bridge the gap between qualitative textual knowledge and quantitative inference, empowering materials scientists to leverage the contextual and empirical spectrum of limited available data for accelerated materials innovation and discovery, with potential applications extending to other material classes beyond alloys.

# Results

## ElementBERT Training and Embedding Harvesting

To enable an effective representation of chemical elements for materials discovery, ElementBERT is designed as a domain-specific language model that captures contextual relationships between elements in alloy-related literature. By leveraging a structured training and embedding extraction workflow, ElementBERT generates high-quality semantic embeddings that serve as universal elemental descriptors. As shown in Figure 1b, ElementBERT is trained on over 1.29 million abstracts from alloy-related scientific literature. The process begins with (i) collecting and preprocessing the abstracts to form an AI-ready corpus. Next, tokenization breaks text into smaller units, like words or sub-words, using a pre-trained tokenizer that follows the BERT training workflow [41, 42]. The training process (ii) employs the Masked Language Modeling (MLM) procedure of BERT, wherein 15% of the tokens are randomly selected for masking. The tokenized inputs are enriched with token and positional encodings before being processed through a stack of N = 12 transformer encoder layers. Each encoder layer consists of n = 6 self-attention heads, residual connections, layer normalization, and a feed-forward network following the DeBERTa-v3_xsmall configuration (see Methods for details) [43]. The model learns to predict the original tokens, with the loss calculated as the cross-entropy between the predicted and true tokens, guiding parameter updates.

Once the ElementBERT encoder (iii) is pre-trained, it facilitates encoding tasks to harvest the chemical element embeddings. After tokenizing individual chemical element symbols (iv), for example "Ti". Semantic embeddings of elements are constructed by encoding the tokenized elements using the pre-trained ElementBERT model, specifically by extracting the 384-dimensional hidden state of the [CLS] token (a special token inserted by the BERT model prepended to each input sequence that serves as an aggregate representation) from the final encoder layer. The resulting embeddings (v) capture contextual information derived from alloy-specific literature, providing a robust representation of chemical elements that can be applied to various downstream applications.

**a** Generating and using chemical element embeddings

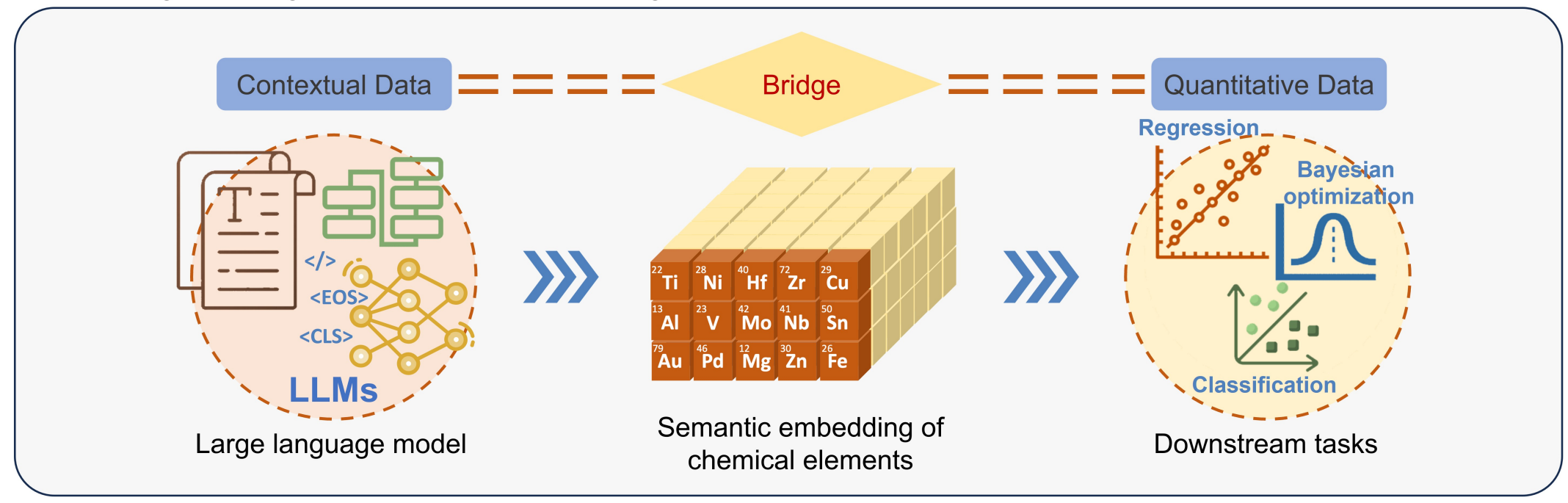

**b** Training and embedding harvesting

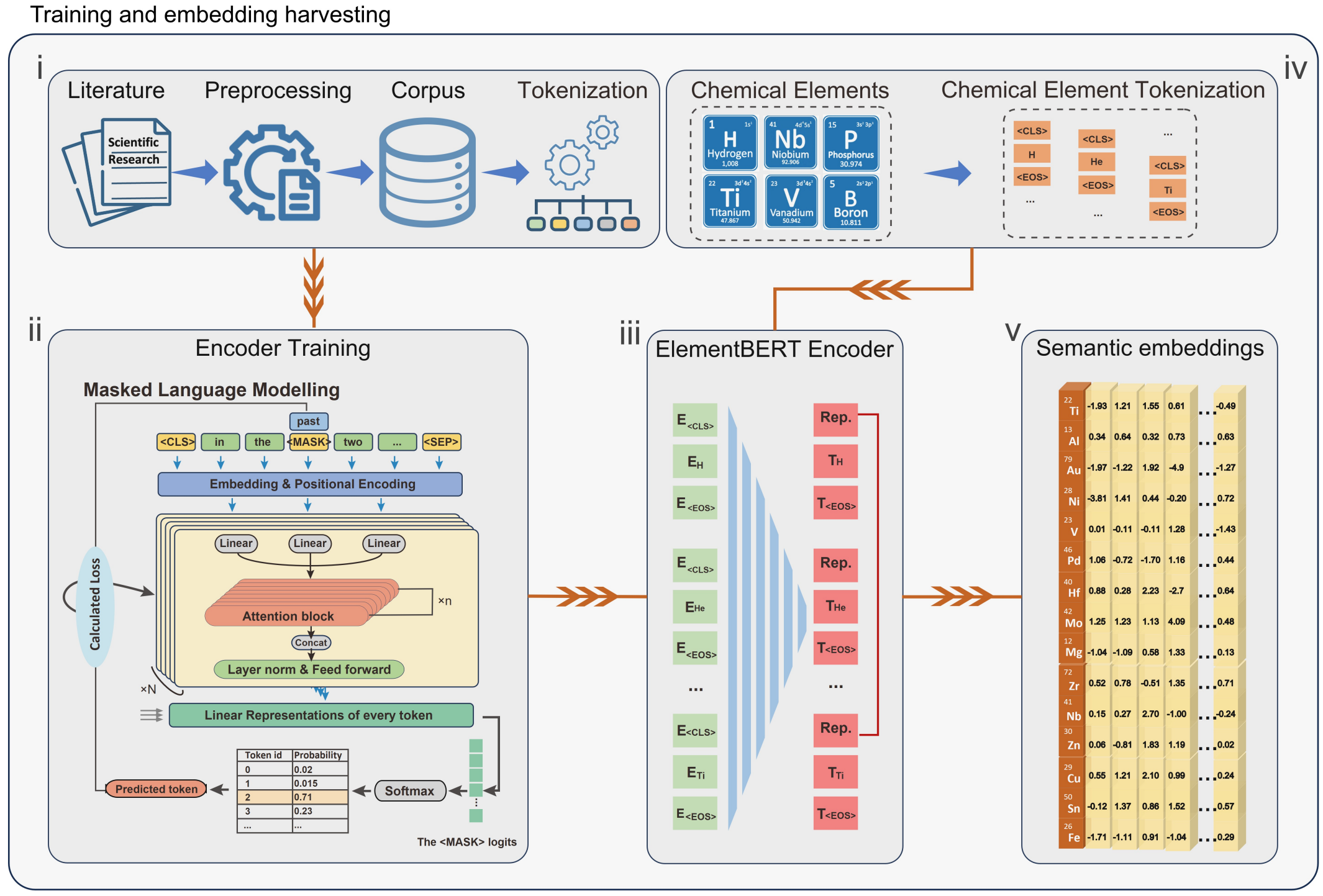

**Figure 1**: **a.** Pipeline for generating semantic embeddings of chemical elements using a large language model, with applications to downstream tasks. **b.** Flowchart of the ElementBERT training and embedding harvesting process. Scientific literature is first compiled into a corpus and tokenized (i) as input for training the ElementBERT encoder (ii). After pre-training, tokenized chemical elements (iv) are passed through the pre-trained ElementBERT model (iii) to generate element embeddings (v), which are then used as semantic representations for chemical elements in downstream applications.

## Materials Inference

### Classification and Regression

We first evaluated the predictive performance of regression and classification models on various properties of shape memory alloys (SMA), Ti alloys, and high-entropy alloys (HEA) using ElementBERT-derived embeddings compared to empirical descriptors of chemical elements (Supplementary information Table S1 and S8). This evaluation demonstrates the utility of ElementBERT embeddings in capturing intrinsic relationships between alloy compositions and their properties.

The experimental datasets, compiled from literature and laboratory experiments, include 295 SMA samples, 603 Ti alloy samples, and 501 HEA samples. Each dataset contains detailed information on the composition, process parameters, and corresponding properties, all of which are publicly available as detailed in our previous work [44]. To ensure robust evaluation, 100 experimental data points were randomly selected from each dataset, followed by feature selection using genetic algorithms. This process was applied separately to traditional elemental descriptors and ElementBERT embeddings to identify optimal feature subsets that minimized the 10-fold cross-validation mean absolute error (MAE). The evaluation process was repeated multiple times to generate MAE distributions.

Figure 2 illustrates the enhanced performance of regression and classification models using semantic embeddings from ElementBERT. Figure 2a compares the 10-fold cross-validated MAE for properties: phase transformation temperatures ($M_p$, $A_p$) and transformation enthalpy ($\Delta H$) for SMAs; yield strength ($\sigma_s$), ultimate tensile strength ($\sigma_s$), and Vickers hardness (VH) for Ti alloys; and $\sigma_s$, $\sigma_s$, and elongation (EL) for HEAs. Each point represents the average of eight parallel tests across varying feature counts. Traditional elemental descriptors (e.g., electronegativity, atomic radius) are indicated by blue lines, while red lines represent ElementBERT-derived features. The shaded regions represent standard deviations of parallel tests. Across nearly all properties and feature counts in the regression models, ElementBERT embeddings consistently achieve lower prediction errors.

For instance, in the case of the martensite phase transformation temperature ($M_p$) of SMA, the reduction in MAE between ElementBERT embeddings and traditional elemental descriptors is minimal when only a few features are selected. However, as the number of features increases, the gap widens, reaching its maximum (over 23%; see the corresponding panel in Figure 2a) at eight selected features. As the number of features increases, the amount of input information increases, which leads to a better model. We observed a more significant decrease in ElementBERT's embeddings. The more obvious decline trend indicates that these semantic embeddings may contain information that does not appear in empirical elemental descriptors. A similar trend is observed in Ti alloys and HEAs, where the MAE for yield strength and hardness decreases by approximately 20% when eight features are used. These findings demonstrate that ElementBERT embeddings are more suitable for alloy composition design, as these embeddings are likely to contain complex relationships between alloy compositions and their properties hidden in the contextual descriptions.

Figure 2b explores classification tasks of varying complexity for phase identification in materials. Specifically, we conducted a binary classification task to distinguish between solid solution (SS) and non-solid solution (NSS) phases in HEAs. As the number of selected features increases, the F1 score improves, reaching nearly 0.9 for ElementBERT embeddings at a feature count of eight. This result highlights the effectiveness of ElementBERT-derived embeddings in enhancing the model's ability to distinguish between SS and NSS phases. While the predictive performance of models using traditional features (blue line) and ElementBERT-derived embeddings (red line) is relatively close when only one or two features are selected, the advantage of ElementBERT embeddings becomes increasingly pronounced as the number of features grows. This further underscores the potential of BERT embeddings in tackling complex material classification tasks.

We also conducted a ternary classification task to distinguish among three phase forms: face-centered cubic (FCC), body-centered cubic (BCC), and a mixed FCC-BCC structure. For SMA materials, we further extended our analysis to a quaternary classification task to distinguish between five phase forms: B19'-B2, B19'-B19-B2, B19'-R-B2, B19-B2, and R-B2. In both HEA and SMA classification tasks, ElementBERT embeddings consistently outperformed traditional descriptors, demonstrating their superior ability to encode meaningful phase-related information and improve classification accuracy.

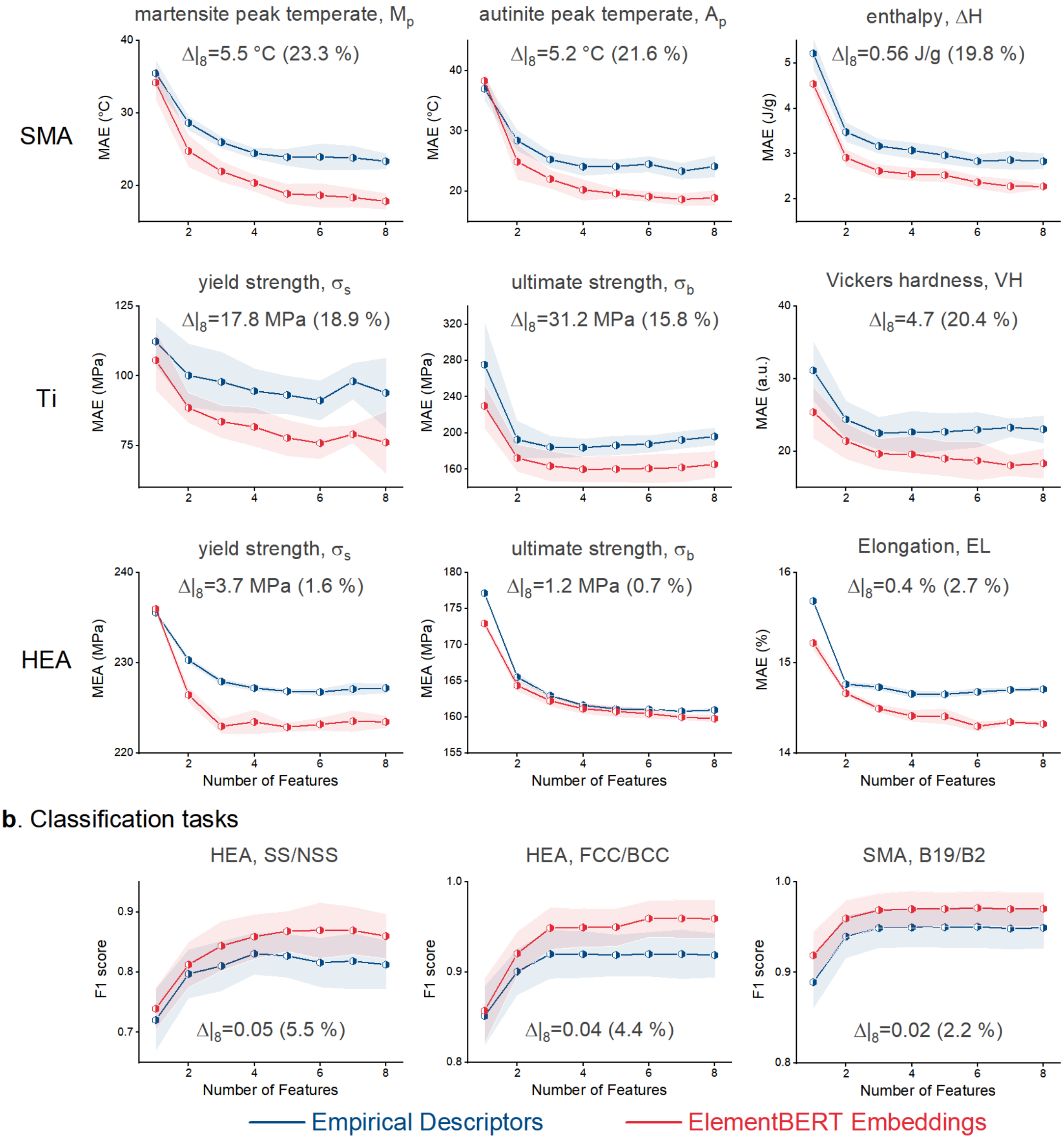

**Figure 2.** Comparison of model performance using ElementBERT-derived features versus empirical features for **(a)** prediction and **(b)** classification of material properties. The 10-fold MAE plots for SMA, Ti alloys, and HEA show performance as a function of the number of selected features (1-8) across extensive parallel tests. Blue lines indicate model performance using traditional empirical features (e.g., electronegativity, atomic radius), while red lines represent ElementBERT-derived material features. Examined properties include phase transformation temperatures ($M_p$, $A_p$), transformation enthalpy ($\Delta H$), yield strength ($\sigma_s$), ultimate tensile strength ($\sigma_b$), Vickers hardness (VH), and elongation (EL). Classification tasks include binary classification of Solid Solution (SS) vs. Non-Solid Solution (NSS), ternary classification of phase forms (Face-Centered Cubic (FCC), Body-Centered Cubic (BCC), and FCC-BCC mixed), and quaternary classification of SMA phases (B19'-B2, B19'-B19-B2, B19'-R-B2, B19-B2, and R-B2). ElementBERT-derived features consistently yield lower prediction errors across nearly all properties and feature counts, highlighting their superior ability to capture intrinsic relationships between alloy composition and properties. Shaded areas represent standard deviation across parallel tests.

### Bayesian optimization

Building on the demonstrated predictive power of ElementBERT-derived embeddings, we now examine whether integrating these embeddings into a Bayesian optimization (BO) framework can more effectively guide the search toward optimal regions in the materials property space. As BO has been proven to be a powerful tool for efficiently exploring complex design spaces in alloy development, we evaluated the effectiveness of ElementBERT-derived embeddings in BO-based materials optimization by comparing them to traditional empirical elemental descriptors across different alloy systems (see Methods for details). For each system, we defined a figure of merit (FOM) consolidating multiple target properties into a single optimization objective. For example, in the HEA system, the FOM was computed as a normalized weighted sum of yield strength, ultimate tensile strength, and elongation. A trained neural network, utilizing experimental data served as the ground truth for predicting FOM values. We used Gaussian Process Regression (GPR) to model the composition-FOM relationship and employed Expected Improvement (EI) as the acquisition function to iteratively suggest new compositions. To ensure robust comparisons, multiple parallel BO trajectories with different random seeds ensured statistical reliability (see Methods for details).

Figures 3a–3c illustrate the evolution of the best-so-far FOM across BO iterations for SMA, Ti alloys, and HEA, respectively. In Figure 3a, the red solid lines represent the mean value of best-so-far FOM values using ElementBERT-derived embeddings across parallel BO runs, while the blue solid lines indicate results using empirical descriptors. The shaded regions represent standard deviations of parallel tests. The inset violin plots display the distribution of final FOM values after all iterations, with each point corresponding to an independent BO run. In the SMA optimization environment, the BO curve obtained using ElementBERT embeddings demonstrates significant improvements in the FOM ($p < 0.05$, unpaired t-test). Similarly, results for HEA and Ti alloys indicate that ElementBERT-derived features yield statistically significant enhancements over empirical descriptors in HEA systems ($p < 0.05$, Figure 3b) while achieving comparable performance in the Ti alloy system ($p = 0.799$, Figure 3c). Notably, the performance gap for BO is more pronounced in the case of HEA, where the use of empirical descriptors approaches saturation in later BO iterations, whereas BO utilizing ElementBERT embeddings continues to show improvement.

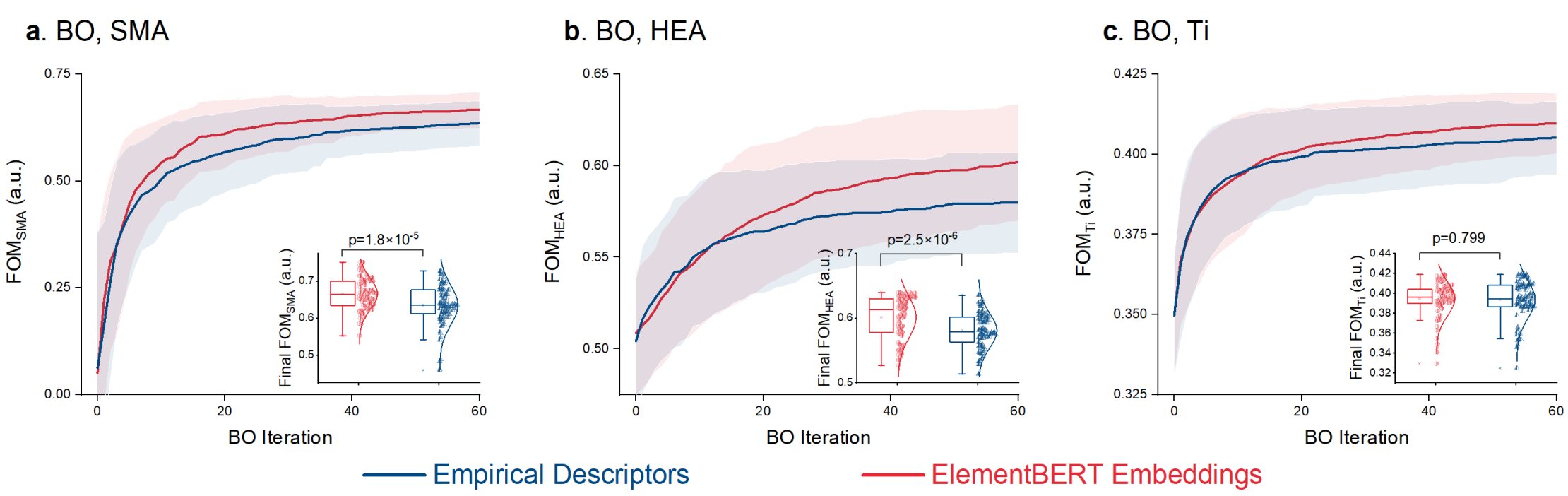

**Figure 3**. Comparison of Bayesian optimization (BO) performance using ElementBERT-derived element embeddings and empirical elemental features across different alloy systems. **(a-c)** Evolution of the best-so-far figure of merit (FOM) during BO iterations for SMA, Ti alloys, and HEA, respectively. The main plots display the mean (solid lines) and standard deviation (shaded areas) of the best FOM values across 96 parallel BO runs. Inset violin plots show the distribution of final FOM values after 60 iterations, with each point representing an independent BO run. ElementBERT features demonstrate statistically significant improvements over empirical features for SMA and HEA systems ($p < 0.01$, unpaired t-test) while achieving comparable performance in the Ti alloy system ($p = 0.799$).

## Prospective validation reveals novel Ti alloys

To assess the generalization capability of ElementBERT embeddings beyond the training corpus, we conducted a systematic evaluation using newly reported titanium alloy data from 2025. These publications were excluded from the ElementBERT corpus up to 2024, and their associated data were not used in the training or evaluation of machine learning models. The experimental workflow, including random sampling around the newly reported compositions, is summarized in Figure 4a. Feature subsets derived from prior regression tasks were used to train machine learning models, after which we collected recently reported titanium alloy compositions and their yield strengths synthesized through experimental methods, ensuring these compositions were entirely absent from both the training corpus and prior experimental datasets. The trained models were then employed to predict the yield strength of these newly reported compositions. The compositional distributions were visualized via t-SNE, and Figure 4b presents a comparison between the spaces defined by ElementBERT embeddings (i) and empirical descriptors (ii). Sampling points predicted to exhibit yield strengths below 1100 MPa were excluded from the plots for clarity. The RF model trained with ElementBERT embeddings identified a substantially larger number of high-performance compositions (blue points) than the model based on empirical descriptors, demonstrating the superior ability of ElementBERT embeddings to capture compositional–property relationships and highlighting their potential to accelerate alloy design in unexplored chemical spaces.

As shown in Figure 4c, we evaluate the performance of 22 widely-used machine learning models on an external 2025 dataset to demonstrate the effectiveness of ElementBERT embeddings in the development of novel alloys (see supplementary information Table S7). Across all models tested, models incorporating ElementBERT embeddings consistently outperformed those based on empirical descriptors. The comparison of median mean errors further substantiates this advantage, confirming that ElementBERT embeddings achieve systematically lower prediction errors.

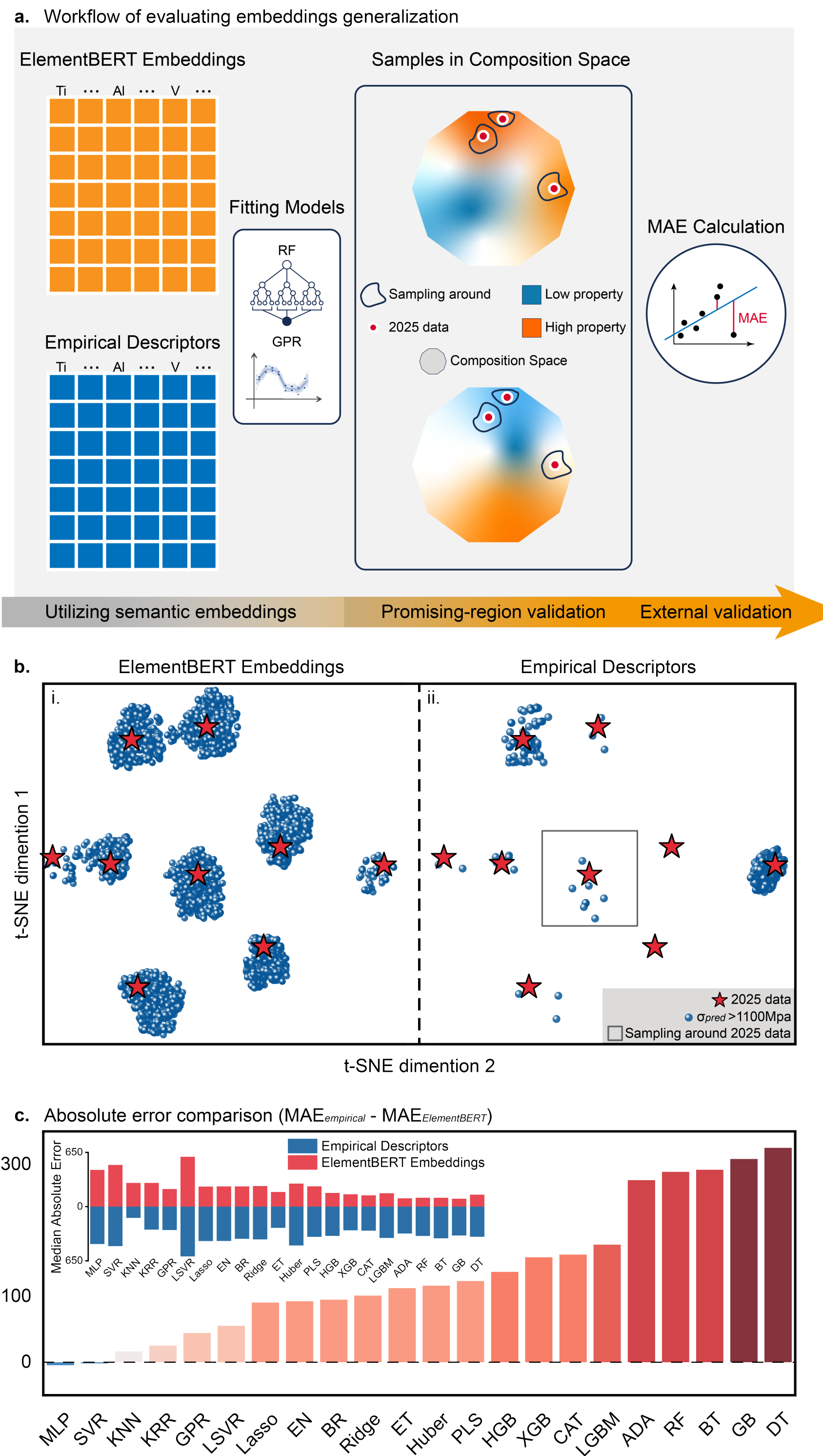

**Figure 4**. Generalization performance of ElementBERT embeddings on newly reported titanium alloy compositions. a. Experimental workflow for evaluating generalization capabilities using newly published titanium alloy data from 2025[45-53], showing the process from Utilizing semantic embeddings and empirical features to do machine learning model fitting to external validations. b. t-SNE visualizations of sampling compositions on two different RF

models based on ElementBERT embeddings (i) and empirical descriptors (ii). The greater amount of sampling points demonstrate that ElementBERT embeddings identify significantly more promising compositions in unexplored chemical spaces. c. MAE improvement across machine-learning algorithms. The main panel reports $MAE_{empirical} - MAE_{ElementBERT}$; higher values indicate larger gains from ElementBERT over empirical descriptors. The inset summarizes median MAE for each method (lower is better).

These results highlight the versatility and effectiveness of ElementBERT embeddings across diverse applications, from handling complex optimization landscapes to generalizing beyond training data for novel alloy discovery. The elemental embeddings provide a compressed representation of contextual elemental information, encoding high-dimensional chemical knowledge into dense vector embeddings that prove robust across different machine learning frameworks and compositional spaces. Through feature selection, we effectively distilled this encoded information to identify the most crucial features, akin to knowledge distillation [54], from the comprehensive elemental embedding space. This process not only reduces dimensionality but also preserves the most relevant chemical insights while enabling generalization to emerging alloy chemistries. These findings underscore the transformative potential of NLP techniques in extracting, encoding, and concentrating domain-specific knowledge for accelerated materials discovery, establishing a foundation for exploring previously inaccessible compositional spaces.

## Discussion

The BERT model derives insights from textual data utilizing a transformer-based architecture [55] that processes text in a bidirectional manner, thereby comprehensively capturing contextual nuances through specialized self-supervised objectives. Specifically, in our study, we employed masked language modelling (MLM), a technique wherein a certain proportion of tokens in each input sequence is randomly obscured ('masked'). The model is subsequently trained to accurately predict the original tokens by leveraging their contextual cues. This iterative process of correctly identifying masked tokens across an extensive corpus of scientific literature compels the model to develop sophisticated contextual embeddings. These embeddings encapsulate the intrinsic relationships within the text, thereby enabling the model to reveal semantic scientific knowledge and thus enhancing its utility in downstream machine learning tasks by providing richer, more informative descriptors that improve predictive accuracy.

Building on the demonstrated effectiveness of semantic embeddings in predictive modeling and optimization, we further analyze their advantage by quantitatively evaluating the similarity and divergence between semantic embeddings and empirical descriptors. Figure 5a presents a heatmap illustrating the Pearson correlation coefficients between widely used empirical elemental descriptors for SMAs, Ti alloys, and HEAs and the ElementBERT embeddings. The heatmap's horizontal axis categorizes the top 90 descriptors based on their similarity within ElementBERT embeddings, while the vertical axis enumerates 30 empirical elemental descriptors. The intensity of the red coloration corresponds to higher Pearson correlation values, effectively visualizing these relationships. The densely correlated areas in dark red on the left indicate that ElementBERT effectively captures empirical physical information, while the more sparsely correlated regions on the right suggest that ElementBERT embeddings encode additional semantic dimensions. This 'orthogonal' information captures high-dimensional relational knowledge—such as the collective experimental consensus on phase stability and synergistic elemental effects—that static physical descriptors inherently miss. By effectively 'digitizing' these complex scientific narratives, the model provides a more holistic and context-aware feature space.

Figures 5b, 5c, and 5d provide detailed correlation analyses for SMAs, Ti alloys, and HEAs, respectively, offering insights into how ElementBERT embeddings relate to empirical descriptors. Taking Figure 5b as an example for HEA, the horizontal axis represents the top 10 ElementBERT embeddings, selected based on their frequency of occurrence, while the vertical axis features the 10 most frequently occurring empirical descriptors, both ordered from highest to lowest frequency. This visualization reveals that semantic embeddings effectively capture critical empirical information, reinforcing

their suitability for feature-specific tasks. For instance, semantic embedding No. 83 exhibits low Pearson correlation coefficients with high-frequency empirical descriptors, indicating that it captures unique information not present in empirical data. This additional knowledge contributes to enhanced model performance. A similar pattern is observed in Ti alloys and SMAs (Figure 5c and 5d).

Because semantic embeddings encode richer information, they distinguish elements more effectively. As a result, elements are more broadly distributed in dimensionality reduction visualizations, deviating from the regular periodicity seen in empirical descriptors. Figure 5e provides further insights into elemental representations using Multidimensional Scaling (MDS) to project cosine similarity measures into two dimensions [56]. These projections demonstrate that empirical feature distributions tend to be more linear and exhibit less variability, limiting their ability to differentiate elements effectively. In contrast, ElementBERT embeddings are more dispersed, offering a broader and more diverse representation of elemental information. This dispersion underscores ElementBERT's superior ability to capture nuanced relationships.

To further assess the effectiveness of ElementBERT-derived embeddings in optimization tasks, we compare their performance with traditional empirical descriptors within BO. As shown in Figure 5f, under identical conditions for random exploration, BO with ElementBERT embeddings enables a more efficient search, expanding exploration beyond the shaded region designated for random sampling. Furthermore, the color variations in this figure indicate that the compositional performance achieved using ElementBERT embeddings is superior to that obtained with empirical descriptors. These results highlight that the broader exploration enabled by ElementBERT embeddings enhances search efficiency and leads to superior optimization outcomes, reinforcing their effectiveness in navigating complex compositional spaces.

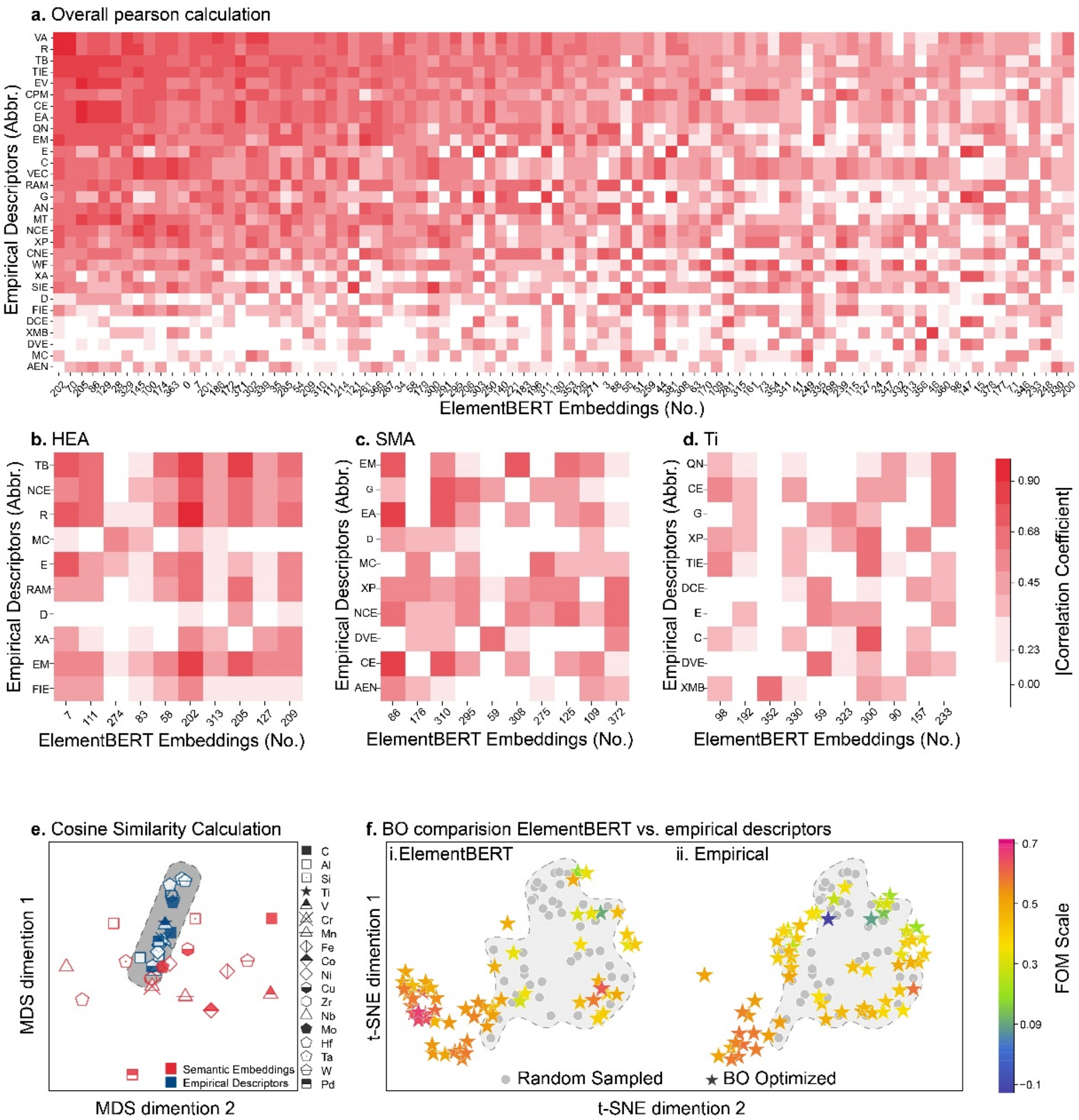

**Figure 5.** Analysis of empirical descriptors and ElementBERT-derived element embeddings. **a.** Heatmap of Pearson correlation coefficients between the most frequently selected features across SMA, Ti-based alloys, and HEA, highlighting relationships between empirical descriptors and ElementBERT embeddings. **b-d.** Correlation heatmaps for HEA **(b)**, SMA **(c)**, and Ti alloy **(d)**, show that ElementBERT embeddings capture not only information similar to empirical descriptors but also additional, weakly correlated knowledge, contributing to improved model performance. **e.** Multidimensional Scaling (MDS) projections comparing cosine similarity distributions of empirical and ElementBERT-derived features. The broader dispersion of ElementBERT embeddings reflects a more diverse and informative elemental representation. **f.** t-SNE visualizations of elemental compositions optimized using BO with ElementBERT-derived embeddings **(i)** and empirical descriptors **(ii)**. The results demonstrate that ElementBERT embeddings facilitate broader exploration of the compositional space, leading to superior optimization outcomes by effectively capturing hidden chemical relationships.

Our research demonstrates that ElementBERT embeddings outperform traditional descriptors and further enhance performance in downstream BO tasks. Additionally, these semantic embeddings exhibit robustness across a range of machine-learning models. Figure 6a presents a comparison of predictive performance across various foundational machine learning models—including Gaussian Process Regression (GPR), Multi-Layer Perceptron (MLP), Random Forest (RF), Support Vector Regression (SVR), and Extreme Gradient Boosting (XGB)—utilizing ElementBERT-derived embeddings. The x-axis represents the 10-fold cross-validation mean absolute error (MAE), with p-values less than 0.01, indicating statistically significant correlations. Across all tested algorithms, ElementBERT consistently surpasses traditional empirical descriptors, demonstrating its adaptability and effectiveness in capturing intricate material-property relationships.

Our findings emphasize the importance of developing domain-specific BERT models to tailor the generated semantic embeddings rather than relying solely on universal large language models. Figure 6b systematically evaluates the influence of different BERT models on predictive performance while keeping all other parameters constant. Organized from left to right by increasing domain specificity, the 10-fold cross-validation MAE distributions for materials property prediction are shown for descriptors derived from raw compositions, empirical features, and embeddings from multiple BERT variants, including DeBERTa-v3_xsmall [43], BERT-base [42], SciBERT [57], MatSciBERT [58], and ElementBERT. The horizontal dashed line in Figure 6b marks the baseline MAE of 24.68 K, achieved using empirical descriptors. As the most domain-specific model, ElementBERT consistently outperforms both general-purpose and other domain-specific BERT architectures, achieving higher predictive accuracy while utilizing fewer computational resources. This advantage is further highlighted by systematic improvements observed in both violin and box plots. The inset in Figure 6b provides a summary of average MAE values across extensive evaluations, reinforcing ElementBERT's superior predictive capabilities over both universal and domain-adapted BERT models.

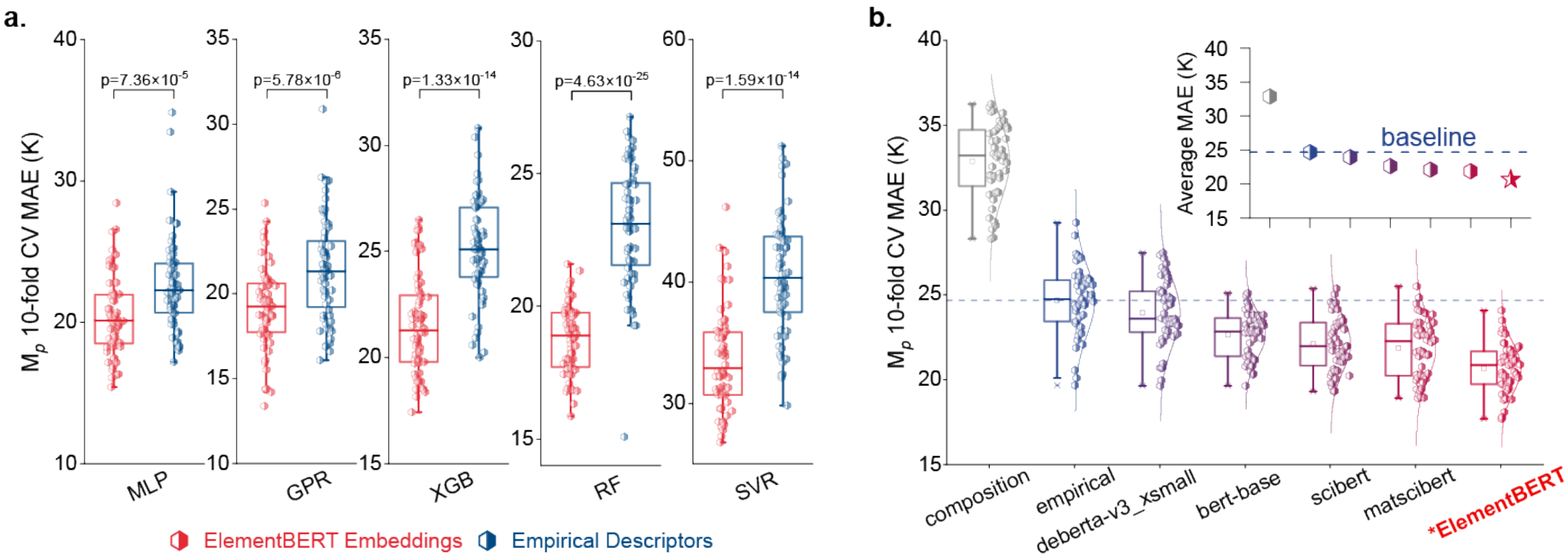

**Figure 6.** Performance comparison of various machine learning models and BERT architectures for materials property prediction. **a.** 10-fold cross-validation MAE distributions for different base models, including Gaussian Process Regression (GPR), Multi-Layer Perceptron (MLP), Random Forest (RF), Support Vector Regression (SVR), and Extreme Gradient Boosting (XGB). Box plots show that ElementBERT embeddings consistently outperform traditional empirical descriptors across all models, demonstrating their adaptability and effectiveness in capturing intricate material-property relationships. **b.** Influence of different BERT architectures on predictive performance for $M_p$ prediction using a RF model. The figure presents 10-fold cross-validation MAE distributions for various input representations, including raw compositions, empirical descriptor-based features, and embeddings from different BERT variants (DeBERTa-v3_xsmall, BERT-base, SciBERT, MatSciBERT, and ElementBERT). The horizontal

dashed line represents the baseline MAE (24.68 K) achieved with empirical descriptors. As the most domain-specific model, ElementBERT consistently outperforms general-purpose and other domain-specific BERT models while requiring fewer computational resources. Violin and box plots systematically illustrate improvements, and the inset summarizes average MAE values across extensive evaluations, highlighting ElementBERT's superior predictive capability.

## Perspective

This study introduces ElementBERT, a domain-specific BERT model trained on alloy-related literature to generate semantic embeddings for chemical elements. These embeddings capture contextual and relational knowledge from scientific texts, enabling significant advancements in predictive modeling, optimization, and materials discovery. ElementBERT consistently outperforms traditional empirical descriptors and general-purpose BERT variants while also requiring fewer computational resources and less computing time. It provides robust and versatile features for regression, classification, and Bayesian optimization across diverse alloy systems, including shape memory alloys, Ti alloys, and high-entropy alloys. Overall, ElementBERT bridges literature-derived linguistic knowledge with quantitative inference, improving both the accuracy and efficiency of materials property prediction and compositional exploration.

The same framework can be extended to generate semantic embeddings for other material's building blocks, such as microstructural features for alloys and crystal structures for non-metal inorganic materials. In particular, this approach is well-suited for capturing semantic embeddings of effective drug targets, active sites, and functional groups, which play a critical role in drug discovery, catalysis, and molecular design. As the framework is dynamic, its embeddings will continuously evolve with the incorporation of larger and more diverse text corpus, leading to progressively refined representations. For task-specific applications, such as catalyst design or photovoltaics, a more targeted, domain-specific corpus can be used to fine-tune large language models, improving their applicability to specialized fields. Given the demonstrated effectiveness of ElementBERT embeddings, we propose the establishment of an open benchmark hub where researchers can contribute and compare semantic embeddings for materials building blocks from various BERT models, facilitating the identification of task-optimized embeddings for mechanical properties, organic functional groups, quantum yield, and other material-related characteristics.

Despite the encouraging results, this study has several limitations that warrant consideration. The pretraining corpus was confined to English-language abstracts available up to 2024. While these abstracts provide a 'distilled' consensus on element-property relationships, they often lack the granular detail of specific processing histories and complex microstructural evolutions found in full-text articles. Consequently, the current embeddings are more sensitive to compositional trends than to fine-grained, processing-dependent structural variations. Further, the evaluation framework emphasized regression and conventional classification tasks, with external validation restricted to recently reported titanium alloys. Extensive experimental validation of Bayesian optimization results will be required to demonstrate broader applicability.

The reliability of ElementBERT embeddings is intrinsically linked to the frequency of elemental mentions in the literature. Although the model covers 119 elements, the corpus naturally exhibits a highly skewed distribution, where ubiquitous alloying elements (e.g., Fe, Ni, Al, Ti) are discussed far more frequently than rare or radioactive ones. Rather than a technical limitation, this distribution represents the actual landscape of materials research. We deliberately avoided artificial corpus balancing during pretraining, as our primary objective was to digitize the ‘latent consensus’—preserving the relative importance and contextual weight of each element as defined by decades of scientific focus. As a result, users should expect the highest semantic fidelity for common elements, while embeddings for rare elements may exhibit reduced robustness in downstream tasks due to their limited data footprint.

Taken together, these limitations highlight clear directions for future research. Expanding the corpus to include full texts, patents, and multilingual sources, together with rigorous experimental validation, will strengthen both the robustness and generality of domain-specific semantic embeddings. Furthermore, acknowledging that model representations are sensitive to the training content, future research should explore how to balance 'domain intuition' (inherited research biases) with objective physical laws. Implementing strategic corpus selection or incremental fine-tuning on specialized datasets will be essential to mitigate inherited biases while refining embeddings for niche applications. Such efforts will consolidate their role as a foundation for accelerated and more reliable materials discovery.

# Methods

## Development of ElementBERT

### Text data collection and preprocessing

To develop a comprehensive dataset for pre-training on alloy-related topics, we employed the widely recognized scientific search engine, Web of Science. Our search was specifically configured with "alloy" as the topic, yielding over 1,290,000 relevant publications. Using the bulk export metadata function, we systematically downloaded these abstracts in batches of 1,000, storing them in Excel spreadsheets. Subsequently, Python scripts were employed to extract only the English texts from these abstracts and consolidate them into a single Markdown file. This procedure was carried out in 2024, thereby inherently limiting the dataset to include only those publications available up to that point.

### Architecture of ElementBERT

ElementBERT is a transformer-based encoding model that amalgamates the encoder architecture of BERT with the "xsmall" version of Microsoft's DeBERTa enhancements. Initially, inputs are tokenized and enriched with both token and positional encodings. Subsequently, these inputs are processed through a stack of $N = 12$ transformer encoder layers, each equipped with $n = 6$ self-attention heads and a hidden size of 384. This configuration contrasts with the standard BERT models, which typically feature 12 transformer encoder layers, 12 self-attention heads, and a hidden size of 768. Despite the reduction in the number of attention heads and the smaller hidden size, ElementBERT efficiently employs a bidirectional attention mechanism, adeptly capturing contextual information from both sides of a word to enhance text comprehension. The integration of DeBERTa's disentangled attention and dynamic positional encoding significantly augments ElementBERT's capability to discern intricate semantic relationships and the importance of word positions. The streamlined attention heads and reduced hidden size contribute to ElementBERT's operational efficiency, rendering it particularly suitable for deployment in environments constrained by computational resources yet still capable of capturing substantial and critical information for analyzing semantic data.

### Training of ElementBERT

The pre-training of ElementBERT was conducted using a single Nvidia A100 40G GPU for approximately 255 hours. The training involved a total of 130,000 steps with an effective batch size of 576 (achieved through 8 samples per device and 72 gradient accumulation steps). The model utilized an AdamW optimizer (torch implementation) with a learning rate of $1e^{-4}$, linear warmup over 10,000 steps, and weight decay of $1e^{-2}$. The training leveraged mixed precision (fp16) to enhance computational efficiency [59].

## Benchmark environments

### Feature subset selection

The datasets used in this study were compiled from multiple sources: SMA data was collected from our laboratory's long-term experiments and literature; HEA data was obtained from [60]; and Ti alloy data was primarily gathered through literature collection (see ref [44] for details). The datasets contain comprehensive values about composition and processing parameters, specifically including 10 elements for SMA, 10 elements for HEA, and 11 elements for Ti alloys.

For feature subset selection, we employed genetic algorithms (GA) to identify optimal feature combinations from both ElementBERT embeddings and empirical descriptors. The GA hyperparameters were carefully tuned: population size was set to 192, crossover rate to 0.8, mutation rate to 0.1, and maximum generations to 100. While the search space dimensions varied between empirical features and ElementBERT embeddings (30 versus 384 dimensions), the optimization budget remained consistent. To ensure robust evaluation, we randomly selected 100 experimental data points and performed feature selection to obtain the subset that minimized 10-fold cross-validation MAE (averaged by four parallel rounds). The feature selection process was conducted independently for each target property and alloy system to ensure task-specific descriptor optimization. The base model for evaluation was RF.

Given an initial feature pool $F = \{f_1, f_2, \ldots, f_n\}$, find $S = \arg\max_{S \subseteq F} Q(S)$. Each GA-optimized feature subset was further subjected to 64 iterations of 10-fold cross-validation using different random seeds. This evaluation procedure, starting from the initial random selection of 100 experimental data points, was repeated at least 8 times (8 feature subsets). Importantly, the selected feature subset for each run was fixed throughout the subsequent cross-validation and evaluation phases to prevent data leakage and ensure a fair comparison. The process generated feature subsets containing $f_n$ features (where $f_n$ ranges from 1 to 8), corresponding to individual performance points shown in our evaluation plots. We also observed that despite the substantially larger search space of ElementBERT embeddings, they consistently yielded superior performance.

### Bayesian optimization

To establish a comprehensive evaluation framework, we developed three distinct test environments (see ref [44] for details), each employing neural networks trained on SMA, Ti, and HEA datasets as the ground truth simulator for generating pseudo-experimental predictions during the BO process. These environments target three alloy families with specific optimization objectives: For shape memory alloys (SMAs), we focus on thermal management properties, combining enthalpy change ($\Delta H$), thermal hysteresis ($\Delta T$), and working temperature ($T_w$, defined as the average of martensite and austenite transformation peak temperatures, $M_p$ and $A_p$) into a figure of merit $\mathrm{FOM_{SMA}} = \frac{1}{3}(\frac{\Delta H}{\Delta H_N} + \frac{\Delta T_N}{\Delta T} + T_{wN} - \frac{\Delta T_{wN}}{T_{wN}})$. Where $\Delta H_N$, $\Delta T_N$, and $T_{wN}$ are normalization factors, and $T_{wN}$ represents the deviation from target working temperature.

The titanium alloy and high-entropy alloy (HEA) environments focus on mechanical properties, with their respective figures of merit defined as $\mathrm{FOM_{Ti}} = \frac{1}{3}(\frac{\sigma_Y}{\sigma_{Y_N}} + \frac{\sigma_U}{\sigma_{U_N}} + \frac{v}{v_N})$, and $\mathrm{FOM_{HEA}} = \frac{1}{3}(\frac{\sigma_Y}{\sigma_{Y_N}} + \frac{\sigma_U}{\sigma_{U_N}} + \frac{\acute{U}}{\acute{U}_N})$. Here, $\sigma_Y$, $\sigma_U$, and ν represent yield strength, ultimate strength, and Vickers hardness for Ti alloys, while the HEA environment replaces hardness with elongation (ε) to balance strength and ductility. All normalization factors ($\sigma_{Y_N}$, $\sigma_{U_N}$, $v_N$, and $\acute{U}_N$) are carefully chosen to ensure comparable scales across different properties. These integrated FOMs create sophisticated multi-objective optimization landscapes that reflect practical material development priorities, providing rigorous benchmarks for evaluating our composition optimization methodology.

The BO procedure follows a workflow: Initially, 40 random compositions are sampled from the compositional space to establish a diverse starting point. Feature subset selection is then performed using GA to identify optimal descriptors from a pool of 30 empirical descriptors or 384 ElementBERT embedding vectors by minimizing the 10-fold cross-validation mean absolute error. The selected feature subset was fixed for all subsequent iterations of the BO loop. Based on these selected features, we construct a mapping function (see ref [44] for details) from composition to materials features, followed by training a GPR model (eq. 1) using materials features and property data. This creates a complete prediction pipeline from composition to target properties.

$$f(x_{1:k}) \sim \mathrm{N}(\mu_0(x_{1:k}), \Sigma_0(x_{1:k}, x_{1:k})). \tag{1}$$

Where $\mu_0(x_{1:k})$ represents the evaluation of the mean function at the input points $[x_1, \ldots, x_k]$, and $\Sigma_0(x_{1:k}, x_{1:k})$ is the covariance matrix calculated between these points using a covariance function or kernel.

The acquisition function calculation chain proceeds through several steps: from composition to materials features, then to property prediction, and finally to the acquisition function value. We employ EI (eq. 2) as the acquisition function, utilizing gradient-based optimization to identify compositions that maximize the acquisition function value within the constrained compositional design space:

$Candidate = \arg\max_{c \subseteq C} EI(c)$ where c represents a composition, C is the compositional design space.

$$\mathrm{EI}(c) = [\mu(c) - f^*]\Phi(\frac{\mu(c) - f^*}{\sigma(c)}) + \sigma(c)\phi(\frac{\mu(c) - f^*}{\sigma(c)}). \tag{2}$$

Where $\mu(c)$ is the predicted mean at the composition c. $\sigma(c)$ is the predicted standard deviation at the composition c. $f^*$ is the highest observed function value so far. $\Phi$ and $\phi$ are the cumulative distribution function (CDF) and the probability density function (PDF) of the normal distribution, respectively.

This process suggests the most promising composition for each iteration, continuing for a total of 60 generations to ensure a thorough exploration of the design space. To ensure further statistical robustness, 96 parallel BO trajectories were conducted with different random seeds, each starting from randomly selected initial compositions. This comprehensive setup allowed for the calculation of the mean and standard deviation of the best-so-far FOM values throughout the iterations.

**Downstream tasks: regression and classification models**

The machine learning implementation utilized the scikit-learn framework to evaluate five distinct algorithms: RF, SVR, GPR, XGB, and MLP. For quantitative performance assessment, we focused on two key metrics: MAE for regression tasks, which provides a direct measure of prediction accuracy, and F1-score (eq. 3,4) for classification tasks. These tasks offer a harmonic mean of precision and recall. This dual-metric approach enables comprehensive evaluation of ElementBERT-derived embedding performance across different types of material property prediction tasks. All models were implemented using standard scikit-learn [61] implementations with consistent cross-validation procedures to ensure robust performance comparison.

$$\mathrm{F1} = 2 \times \frac{\mathrm{Precision} \times \mathrm{Recall}}{\mathrm{Precision} + \mathrm{Recall}} \tag{3}$$

Where, $\mathrm{Precision} = \frac{TP}{TP + FP}$, and $\mathrm{Recall} = \frac{TP}{TP + FN}$. True Positives (TP) are correctly predicted positive cases, False Positives (FP) are incorrectly predicted as positive, and False Negatives (FN) are positive cases incorrectly predicted as negative.

Multiclass F1 score (eq. 4) are calculated by weighted F1 score:

$$\mathrm{F1}_{\mathrm{weighted}} = \sum_{i=1}^{N} \left( \frac{|y_i|}{\sum_{j=1}^{N} |y_j|} \right) \times \mathrm{F1}_i \tag{4}$$

Where N is the total number of classes and $|y_i|$ represents the number of samples in the $i-th$ class.

During the machine learning phase, several key points warrant mention. In particular, hyperparameter tuning was conducted using the BayesianOptimization function from the bayesian-optimization library [62] in combination with scikit-learn models, leveraging empirical descriptors to ensure optimal performance. For further implementation details, please refer to the supplement. Despite these efforts, algorithms that utilize semantic embeddings as inputs consistently outperform those that rely solely on empirical features. This highlights the superior efficacy of semantic embeddings in enhancing algorithmic performance.

**External Validation on Recent Literature**

**Newly reported Datasets Collection**

To rigorously evaluate the generalization capabilities of ElementBERT embeddings on completely unseen data, we conducted external validation using titanium alloy compositions and properties reported in newly published literature from 2025. To ensure data quality and relevance, we applied the following inclusion criteria: (1) titanium alloys synthesized through experimental methods; (2) papers with primary focus on composition development and alloy design; (3) peer-reviewed publications. These compositions represent cutting-edge alloy development recognized by the scientific community, providing a stringent test for the generalization capabilities of ElementBERT embeddings beyond the training data temporal boundary.

**Machine learning model Portfolio**

A comprehensive suite of 22 regression models was evaluated, covering all major machine learning approaches for materials property prediction. The models were categorized into distinct algorithmic families to ensure comprehensive coverage of the regression landscape. The full names and descriptions are showed in Table S1.

**Composition sampling**

To enable systematic composition exploration, the continuous composition space was discretized using a uniform grid with step size $\Delta = 0.01$ (1 at.%). This discretization transforms the composition space into integer units where: $x_i = \frac{n_i}{100}$ where $n_i$ is an integer representing the number of units for element $i$, and $\sum_{i=1}^{11} n_i = 100$.

A local sampling strategy was implemented to explore composition neighborhoods around candidate alloys. Starting from each external candidate composition, the algorithm performs discrete random walks within a constrained radius ($r_{\max} = 3$) shown in algorithm 1:

**Algorithm 1** Local Discrete Composition Sampling

**Require:** $\mathbf{C}$: candidate compositions matrix
**Ensure:** $\mathbf{S}$: generated sample compositions

1: Initialize $\mathbf{S} \leftarrow \emptyset$, $\delta \leftarrow 0.01$, $units \leftarrow 100$
2: **for each** $\mathbf{c} \in \mathbf{C}$ **do**
3:     $\mathbf{u}_{start} \leftarrow \text{round}(\mathbf{c} \times units)$
4:     $\mathbf{u}_{start} \leftarrow \text{clip}(\mathbf{u}_{start}, \mathbf{u}_{min}, \mathbf{u}_{max})$
5:     Adjust $\mathbf{u}_{start}$ to satisfy $\sum \mathbf{u}_{start} = units$      ▷ Simplex constraint
6:     **for** $i = 1$ to 1000 **do**
7:         $\mathbf{u} \leftarrow \mathbf{u}_{start}$
8:         $steps \leftarrow \text{random}(1, 3)$
9:         **for** $j = 1$ to $steps$ **do**
10:             $donors \leftarrow \{k : \mathbf{u}[k] > \mathbf{u}_{min}[k]\}$
11:             $receivers \leftarrow \{k : \mathbf{u}[k] < \mathbf{u}_{max}[k]\}$
12:             **if** $donors \neq \emptyset$ and $receivers \neq \emptyset$ **then**
13:                 $d \leftarrow \text{random_choice}(donors)$
14:                 $r \leftarrow \text{random_choice}(receivers)$
15:                 **if** $d \neq r$ **then**
16:                     $\mathbf{u}[d] \leftarrow \mathbf{u}[d] - 1$
17:                     $\mathbf{u}[r] \leftarrow \mathbf{u}[r] + 1$
18:                 **end if**
19:             **end if**
20:         **end for**
21:         $\mathbf{S} \leftarrow \mathbf{S} \cup \{\mathbf{u}/units\}$
22:     **end for**
23: **end for**
24: Remove duplicates from $\mathbf{S}$
25: **return** $\mathbf{S}$

## Code and Data Availability

All code, datasets, and trained models are openly available at Zenodo (https://doi.org/10.5281/zenodo.17402885) and GitHub (https://github.com/diegoxue/ElementBERT). The code is released under the MIT License; datasets are released under CC0 1.0 Universal Public Domain Dedication.

The repository includes: (1) complete source code for ElementBERT pretraining and downstream tasks; (2) the 2025 publications training and test datasets; (3) semantic embeddings and experimental property datasets; (4) comprehensive documentation for full reproducibility. The pre-trained model is also available on HuggingFace at https://huggingface.co/Neuquar/ElementBERT.

Training corpus metadata is derived from Web of Science Core Collection (Clarivate) with proper attribution. The full corpus requires institutional access to Web of Science. For details of the classification datasets, please refer to Zhang et. al. [63], where the corresponding source data is provided.

## Acknowledgment

The authors gratefully acknowledge the support of the National Key Research and Development Program of China (2021YFB3802100), National Natural Science Foundation of China (Grant Nos. 52173228, 52271190 and 524B2165), and National Advanced Rare Metal Materials Technology Innovation Center Project (Program No.2024ZG-GCZX-01(1)-06).

## Competing interests

The authors declare no competing interests.